\documentclass[conference]{IEEEtran}
\IEEEoverridecommandlockouts
\usepackage{cite}
\usepackage{amsmath,amssymb,amsfonts}
\usepackage{algorithmic}
\usepackage{graphicx}
\usepackage{textcomp}
\usepackage{xcolor}

\usepackage{algorithm}
\usepackage{algorithmic}
\usepackage{booktabs}

\usepackage{multirow}
\usepackage{tikz}
\usepackage{enumitem}
\usepackage{placeins}
\usepackage{url}

\usepackage[breakable]{tcolorbox}
\usepackage{capt-of}
\usepackage[T1]{fontenc} 

\newcommand{\ours}{TaxoScale}
\newcommand{\stitle}[1]{\noindent\textup{\textbf{#1}}}
\newcommand{\myNum}[1]{(\emph{#1})}

\def\BibTeX{{\rm B\kern-.05em{\sc i\kern-.025em b}\kern-.08em
    T\kern-.1667em\lower.7ex\hbox{E}\kern-.125emX}}
\begin{document}

\bstctlcite{BSTcontrol}

\title{Security and Privacy Taxonomy Generation from Mobile App Reviews}

 
\author{\IEEEauthorblockN{Moghis Fereidouni\IEEEauthorrefmark{1}, Krishna Upadhyay
\IEEEauthorrefmark{2}, Umar Farooq\IEEEauthorrefmark{2}, A.B. Siddique\IEEEauthorrefmark{1}}
\IEEEauthorblockA{\IEEEauthorrefmark{1}
    University of Kentucky,
    \IEEEauthorrefmark{2} Louisiana State University \\
 Email: 
moghis.fereidouni@uky.edu, kupadh4@lsu.edu, ufarooq@lsu.edu, ab.siddique@uky.edu
}}


\maketitle

\begin{abstract}
Mobile app reviews are a rich, continuously renewing source of how users experience privacy and security, yet existing taxonomies of these concerns are hand-crafted and cannot keep pace with the evolving nature of the data. 
Automating taxonomy construction is the natural response, but scalability is the core challenge: current LLM- and clustering-based methods are developed for scientific corpora of a few thousand documents and do not extend to app review collections numbering in the hundreds of thousands.
We address this gap in two ways. First, we filter app reviews for privacy- and security-related content, yielding a comprehensive corpus of over 600K reviews. Second, we introduce {\ours}, a pipeline that handles taxonomy construction at this scale by extending an expert-defined taxonomy via \emph{Recursive Hierarchical Clustering} and LLM-based node naming.
{\ours} outperforms strong automatic-taxonomy baselines on path, level, coverage, and novelty metrics, and discovers novel branches absent from prior taxonomies. We publicly release our code and data\footnote{\url{https://anonymous.4open.science/r/PrivacyTaxonomy-613D}} to support reproducibility and future research.
\end{abstract}

\section{Introduction}


Mobile apps expose users to a wide range of privacy and security concerns, such as unclear data collection, intrusive permissions, and surveillance behaviors. 
Taxonomies that organize these concerns into a structured hierarchy are a foundational tool for addressing them: they give developers and regulators a shared vocabulary and enable systematic assessment of apps. Mobile app reviews are an exceptionally rich resource for constructing such a taxonomy. They offer one of the largest and fastest-growing records of real privacy and security experiences ever produced. Every day, users describe permissions they find intrusive, data they suspect is being collected, surveillance behaviors they have noticed, and accounts they believe have been compromised. Unlike privacy policies or expert audits, these reviews capture privacy and security as it is experienced, at the scale of entire app ecosystems. They are also unusually responsive to change: as apps ship new features, as platforms revise permission models, and as new threats emerge, the concerns voiced in reviews shift accordingly. This combination (massive scale, authentic user perspective, and continuous renewal) makes app reviews a uniquely valuable resource for extracting privacy and security concerns and organizing them into a taxonomy that evolves with them.

Prior work has recognized this value and mined app reviews for privacy and security signals. Zhang et al. \cite{MiningPrivacyZHANG2025} extracts user privacy concern topics from reviews,  Nema et al. \cite{Security_Privacy_Nema} analyzes user perspectives on mobile app privacy at scale, and Harkous
et al. \cite{Security_Privacy_Harkous} introduces Hark, a deep learning system for discovering and summarizing privacy feedback. Beyond extraction, some of this work organizes the resulting concerns into structured taxonomies. However, these taxonomies are hand-crafted \cite{privacyTaxonomy}. Manual construction carries real consequences. It is expensive, requiring sustained expert effort to build. And it is static; frozen at the moment of creation and unable to adapt as concerns evolve. The very property that makes app reviews valuable (their fast-evolving, ever-renewing nature) is precisely what a hand-crafted taxonomy cannot keep pace with. This tension motivates our work.

\begin{table*}[t]
\caption{Summary statistics of the mobile app review dataset after filtering for English-language reviews with a minimum length of five words.} 
\centering
\resizebox{0.8\textwidth}{!}{%
    \begin{tabular}{lcccc}
    \hline
    \textbf{Feature} & \textbf{AI-related} & \textbf{MobileRec} & \textbf{Teacher Approved} & \textbf{Total} \\
    \hline
    Number of Reviews        & 2.88M   & 14.21M  & 1.54M & 18.63M \\
    Number of Apps           & 292     & 10.17K & 1.12K & 11.22K  \\
    Number of App Categories & 14         & 48     & 18    &48  \\
    \hline
    \end{tabular}%
}
\label{tab:dataset_statistics}
\end{table*}

The natural response is to automate taxonomy construction. In recent years, large language models have risen rapidly and been adopted across a wide range of tasks \cite{achiam2023gpt,touvron2023llama,guo2025deepseek, ESLAMIAN2026132928, shoaeinaeini2026activationsteered, RealTOD_Fereidouni, Shoaeinaeini_IEEE, fereidouni-etal-2024-grounded}, including the automatic construction of taxonomies. A substantial literature now uses LLMs and clustering methods to automatically generate taxonomies \cite{kargupta-etal-2025-taxoadapt, gao2025science, zhu-etal-2025-context, katz-etal-2024-knowledge, hu2024taxonomy, hsu-etal-2024-chime}. However, these methods were developed for scientific and computer science literature, where corpora typically span only a few thousand documents; small enough to fit within an LLM context window or to be processed with modest clustering. App reviews are the opposite regime: large-scale corpora can contain tens of millions of reviews, and even after filtering for privacy- and security-related content, the resulting set may still number in the hundreds of thousands. As a result, these methods do not readily apply at this scale: prompting-based approaches cannot ingest the full label set, and embedding-based clustering methods become intractable at this scale. This is the first problem we address: existing taxonomy generation methods cannot handle large-scale app review corpora.

A second obstacle is the lack of suitable data. No large-scale, broad-coverage open corpus of privacy- and security-filtered app reviews currently exists. The datasets used in prior privacy-focused work on app reviews \cite{Security_Privacy_Nema, Security_Privacy_Harkous, privacyTaxonomy} are not publicly released, and general-purpose app review datasets \cite{MobileRecMaqbool, PPrior_Fereidouni} span every topic without any privacy or security filtering. The few open privacy-focused review resources that do exist are limited in scope: HARPT \cite{kelly2025health} is confined to health-related apps, and the resource released by Zhang et al. \cite{MiningPrivacyZHANG2025} covers only privacy concerns in eleven popular apps. Neither is designed as a general corpus spanning the full range of app categories. This is the second problem we address.

To address these problems, we first build an LLM-based classifier to extract privacy- and security-related reviews from a large multi-source corpus, evaluating ten open-source LLMs across eight prompt templates against a human-annotated benchmark and selecting the strongest configuration to filter all 18.63M reviews down to 601,257 privacy/security-related reviews. We then apply an extractive summarizer to each review, isolating the specific text spans that describe a concern and turning noisy, free-form reviews into clean, concern-level pseudo labels. To organize these pseudo labels into a taxonomy at scale, we introduce {\ours}, a five-stage pipeline that seeds from an existing expert taxonomy and extends it by combining embedding-based clustering with LLM-based node naming. At its core, {\ours} introduces Recursive Hierarchical Clustering, which recursively alternates top-down k-means partitioning with bottom-up Ward agglomerative merging to scale gracefully to hundreds of thousands of labels while remaining grounded in established privacy terminology. {\ours} outperforms strong automatic-taxonomy baselines across path, level, coverage, and novelty metrics, and surfaces concrete novel branches, including an entirely new Network Security branch and a domain-specific Parental Control branch absent from prior taxonomies.

We summarize our contributions as follows:

\begin{itemize}
\item To the best of our knowledge, we are the first to automatically expand a privacy and security taxonomy from over 600K privacy- and security-related mobile app reviews.
\item We introduce {\ours}, a five-stage pipeline that extends an existing expert-defined taxonomy by combining embedding-based clustering with LLM-based node naming, producing finer subcategories and uncovering novel branches. 
\item We open-source the first large-scale collection of privacy- and security-filtered app reviews, together with their extracted concern labels, providing a reusable resource for future research on privacy/security in mobile apps.
\end{itemize}

\section{Dataset Construction}


\subsection{Diverse App Review Aggregation}

We constructed the review dataset by aggregating data from three distinct sources: AI-powered app reviews provided by Chhetri et al. \cite{AIPoweredVinaik}, the MobileRec dataset \cite{MobileRecMaqbool}, and reviews sourced from teacher-approved applications on Google Play. The detailed statistics of these three datasets can be found in Table \ref{tab:dataset_statistics}.


\subsection{Privacy/Security Filter}

\begin{table*}[t]
\caption{F1 scores across all model and prompt template combinations.}
\centering
\resizebox{\textwidth}{!}{%
\begin{tabular}{l|c|cc|ccc|cc|cc}
\toprule
\multirow{2}{*}{\textbf{Prompt Template}} & 
\textbf{Mistral} & 
\multicolumn{2}{c|}{\textbf{Mixtral}} & 
\multicolumn{3}{c|}{\textbf{Gemma-3}} & 
\multicolumn{2}{c|}{\textbf{Llama}} & 
\multicolumn{2}{c}{\textbf{DeepSeek-R1-Distill}} \\
\cmidrule(lr){2-2}\cmidrule(lr){3-4}\cmidrule(lr){5-7}\cmidrule(lr){8-9}\cmidrule(lr){10-11}

 & 
\textbf{7b} & 
\textbf{8x7b} & \textbf{8x22b} & 
\textbf{4b} & \textbf{12b} & \textbf{27b} & 
\textbf{3.1-8b} & \textbf{3.3-70b} & 
\textbf{llama-70b} & \textbf{qwen-32b} \\
\midrule
Zero-shot (naive)             & 0.9634 & 0.9434 & 0.9300 & 0.8163 & 0.9434 & 0.9098 & 0.9467 & 0.9600 & 0.9567 & 0.9600 \\
Zero-shot (rules)             & 0.9600 & 0.8919 & 0.9434 & 0.8019 & 0.9667 & 0.9367 & 0.9300 & \textbf{0.9700} & 0.9231 & 0.9500 \\
Zero-shot (permissions)       & 0.9667 & 0.9267 & 0.9401 & 0.5400 & 0.9567 & 0.9333 & 0.9401 & 0.9634 & 0.9400 & 0.9467 \\
Zero-shot (permissions+rules) & 0.9398 & 0.9128 & 0.9534 & 0.5890 & 0.9634 & 0.9467 & 0.9367 & 0.9667 & 0.9196 & 0.9299 \\
\midrule
Few-shot (naive)              & 0.9434 & 0.9501 & 0.9333 & 0.7835 & 0.9500 & 0.9400 & 0.9031 & 0.9500 & 0.9501 & 0.9500 \\
Few-shot (rules)              & 0.9534 & 0.9400 & 0.9334 & 0.8091 & 0.9534 & 0.9534 & 0.9267 & 0.9634 & 0.9467 & 0.9433 \\
Few-shot (permissions)        & 0.9467 & 0.9434 & 0.9434 & 0.5551 & 0.9467 & 0.9467 & 0.8349 & 0.9534 & 0.9534 & 0.9600 \\
Few-shot (permissions+rules)  & 0.9567 & 0.9500 & 0.9401 & 0.6303 & 0.9500 & 0.9534 & 0.8797 & \textbf{0.9700} & 0.9400 & 0.9634 \\
\bottomrule
\end{tabular}%
}
\label{tab:classification_results}
\end{table*}

Since mobile app reviews frequently include feedback on a wide range of topics, it is necessary to first identify and extract only those reviews that specifically discuss privacy- and security-related concerns to meaningfully analyze user perspectives. To establish a reliable evaluation benchmark for this filtering step, we manually sampled 300 user reviews from the corpus to construct a high-quality evaluation benchmark for privacy/security relevance classification. Two authors independently annotated each review as either privacy/security-related or non-privacy/security-related, based on whether the review explicitly referenced privacy or security concerns. Agreement between annotators was measured using Cohen's kappa \cite{Cohen1960ACO}, which indicated a substantial agreement ($\kappa$ = 0.87). For the finalized benchmark, a third author acted as a tie-breaker and resolved the label for the reviews where there was a disagreement between the first two authors, resulting in a set with unambiguous ground truth for subsequent model evaluation.

To develop a reliable classifier for identifying privacy- and security-related reviews, we systematically evaluated multiple open-source LLMs across a variety of prompt templates, aiming to determine which model and prompting strategy yield the best classification performance on our benchmark.

\subsubsection{Models}
We evaluated ten open-source LLMs spanning diverse scales and architectures, including dense, Mixture-of-Experts, and distilled variants from the Llama \cite{grattafiori2024llama}, Gemma \cite{team2025gemma}, Mixtral \cite{jiang2024mixtral}, and DeepSeek \cite{guo2025deepseek} families. This selection enables comparisons across model families, sizes, and architectural designs.

\subsubsection{Prompting Strategies}

To systematically assess sensitivity to prompt formulation, we designed eight templates combining three dimensions: zero- versus few-shot prompting, explicit classification rules, and app-permission context.

\stitle{prompting paradigm}: zero-shot, where the model receives only instructions with no examples, versus few-shot, where three labeled examples are included to guide the model's output.

\stitle{classification rules}: some prompts provide an explicit privacy- and security-related topics that should be labeled "Yes" (e.g., surveillance, data leaks, identity theft) as well as categories that should be labeled "No" (e.g., UI complaints, performance issues), while others rely solely on the model's inherent understanding of the task. 

\stitle{app permissions}: Some prompts additionally supply the list of permissions declared by the application, providing a contextual signal that may help the model better interpret ambiguous reviews. 

Combining these three binary dimensions yields the full set of eight templates: zero-shot naive, zero-shot rules, zero-shot permissions, zero-shot permissions+rules, few-shot naive, few-shot rules, few-shot permissions, and few-shot permissions+rules.

\subsubsection{Comparison of Models and Prompting Strategies}
We evaluated every model--prompt combination on the human-annotated benchmark described above. As shown in Table \ref{tab:classification_results}, across all tested models and prompting strategies, Llama-3.3-70B achieved the highest F1 score (0.9700) under two distinct prompt templates: zero-shot (rules) and few-shot (permissions+rules). More broadly, Llama-3.3-70B demonstrated consistently strong performance across all eight templates, being the only model to achieve an F1 score of at least 0.95 in every configuration. Gemma-3-4B, by contrast, showed the weakest and most variable performance, particularly when permissions were included in the prompt, suggesting that smaller models struggle to effectively leverage the additional context. These results establish Llama-3.3-70B as the most reliable model for this classification task among all candidates evaluated.

Regarding the choice of prompt template, both zero-shot (rules) and few-shot (permissions+rules) yielded identical peak performance on our 300-review benchmark. We selected few-shot (permissions+rules) for the final pipeline, as we hypothesize that the combination of few-shot examples, explicit classification rules, and app permission context provides the model with richer signal that is more likely to generalize to the full, noisier corpus. Applying this configuration to the complete dataset of 18.63M reviews yielded 601,257 privacy- and security-related reviews, which form the basis for the subsequent privacy/security summarization and taxonomy generation.

\subsection{Extractive Privacy/Security Summarizer}
\label{extractive_summarizer}

Beyond classifying reviews as privacy/security-related, we require an extractive summarizer capable of precisely identifying the specific spans of text within each review that describe the underlying privacy or security concern.

To evaluate Llama-3.3-70B's suitability before applying it to the full corpus, we constructed a held-out evaluation benchmark of 140 randomly sampled privacy/security-related reviews. Since a single review may contain multiple distinct concerns, annotators were instructed to produce multiple extractive spans per review where appropriate.

To establish ground truth, two authors independently read each review and identified text spans that captured all privacy- and security-related concerns expressed in the reviews. Inter-annotator agreement was then measured using Cohen's kappa \cite{Cohen1960ACO}, yielding a score of $\kappa = $ 0.78, which reflects substantial agreement beyond chance. When the spans selected by the two annotators did not fully overlap, the case was automatically flagged as a disagreement. A third author then reviewed the flagged cases and determined the final spans used as ground truth.

We then ran Llama-3.3-70B on the same 140 reviews and evaluated its outputs against the finalized ground truth using two metrics: token-level F1, which measures word-overlap between the model's extracted spans and the human-annotated spans, and Jaccard similarity, which measures span-level overlap. The model achieved a Jaccard similarity of 0.8511 and a token-level F1 of 0.8187, both indicating strong extractive summarization performance. Satisfied with these results, we applied this pipeline to all 601,257 privacy/security-related reviews to extract the privacy/security concern-describing spans that serve as input to the subsequent taxonomy generation stage. We refer to these extracted spans as pseudo-labels.


\section{Methodology}
\label{sec:method}



\begin{figure*}[h]
    \centering
    \includegraphics[width=\textwidth]{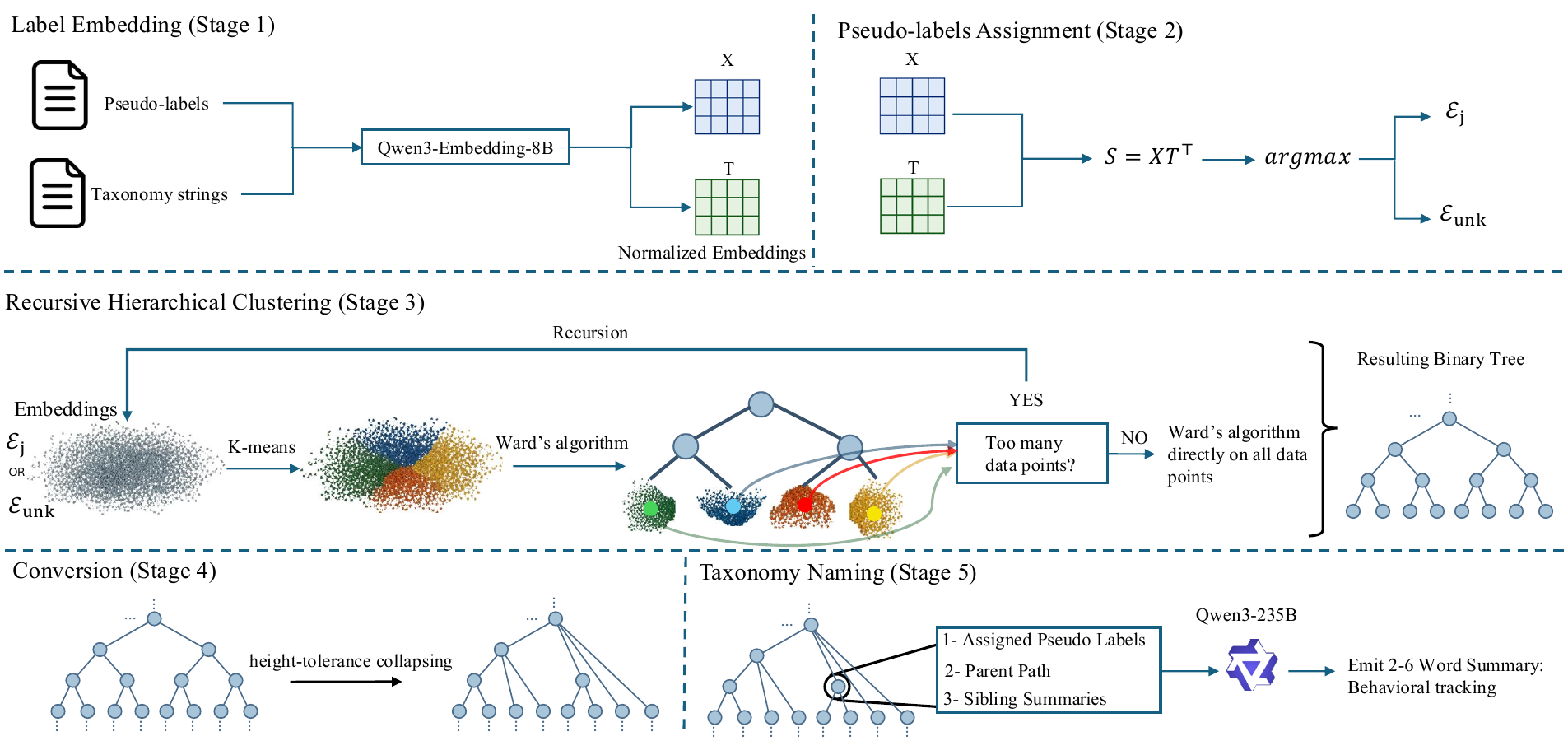}
    \caption{Our five-stage taxonomy construction pipeline: (1) label embedding, (2) pseudo-label assignment, (3) recursive hierarchical clustering, (4) binary-to-multiway tree conversion, and (5) LLM-based node naming (Qwen3-235B).}
    \label{fig:methodology}
    \vspace{-10pt}
\end{figure*}

When a domain-expert taxonomy already exists, building the taxonomy from scratch wastes prior knowledge and may produce categories that diverge from established terminology. 
Our method {\ours}, therefore, seeds the tree with a predefined privacy/security taxonomy and extends it to provide finer subcategories. 
The seed in our experiments is the privacy taxonomy of prior work~\cite{privacyTaxonomy}: a two-level structure of $20$ top-level categories (e.g., \emph{Data Collection}, \emph{Data Sharing}, \emph{Consent}, \emph{Surveillance}, \emph{Advertising}) with $0$--$13$ named subcategories each. The pipeline proceeds in five stages, summarized in Figure~\ref{fig:methodology}.

\subsection{Stage 1: Label Embedding}
\label{sec:method:embed}

Let $L = \{\ell_1, \ldots, \ell_N\}$ denote the set of pseudo-labels generated in Section~\ref{extractive_summarizer}, where $N$ is the total number of pseudo-labels. Each pseudo-label $\ell_i \in L$ is encoded with the Qwen3-Embedding-8B~\cite{zhang2025qwen3} sentence encoder and $\ell_2$-normalized:

\[
    x_i \;=\; \frac{f_{\text{enc}}(\ell_i)}{\lVert f_{\text{enc}}(\ell_i) \rVert_2},
    \qquad
    X \in \mathbb{R}^{N \times d}.
    \label{eq:embed}
\]
Taxonomy nodes are embedded analogously. For each parent--child pair in the predefined taxonomy, we construct two strings: the concatenated form \textit{parent\,--\,child} and the parent name alone. For example, the parent \textit{Selling data} with the child \textit{For advertising purposes} yields two labels: \textit{Selling data--For advertising purposes} and \textit{Selling data}. All resulting strings are passed through the same encoder and $\ell_2$-normalized:
\[
    t_j \;=\; \frac{f_{\text{enc}}(s_j)}{\lVert f_{\text{enc}}(s_j) \rVert_2},
    \qquad
    T \in \mathbb{R}^{M \times d},
    \label{eq:embed_tax}
\]
where $s_j$ denotes the $j$-th taxonomy string and $M$ is the total number of such strings.

\subsection{Stage 2: Pseudo-label Assignment}
\label{sec:method:assign}

After constructing the $\ell_2$-normalized matrices $X$ and $T$, we efficiently compute the cosine similarity between every pseudo-label string and every taxonomy string via matrix multiplication:
\[
    S \;=\; X \, T^{\top} \;\in\; [-1,1]^{N \times M}.
    \label{eq:cosmat}
\]
For each pseudo-label $\ell_i$, the closest taxonomy node is identified as:
\[
    j^{*}_i \;=\; \arg\max_{j}\; S_{i,j},
    \label{eq:argmax}
\]
with corresponding similarity score $s_i = S_{i,\, j^{*}_i}$. Using a threshold $\theta = 0.5$, we collect the embeddings of all confidently assigned pseudo-labels for each taxonomy node $j$:
\[
    \mathcal{E}_j \;=\; \{ x_i \mid j^{*}_i = j,\; s_i \geq \theta \},
    \label{eq:cluster}
\]
where $|\mathcal{E}_j|$ denotes the number of pseudo-labels assigned to node $j$. Pseudo-labels that fall below this threshold are considered unassignable, and their embeddings are collected separately:
\[
    \mathcal{E}_{\text{unk}} \;=\; \{ x_i \mid s_i < \theta \},
    \label{eq:unk}
\]
which are attached directly to the root during taxonomy construction.


\subsection{Stage 3: Recursive Hierarchical Clustering}
\label{sec:method:cluster}

\begin{algorithm}[t]
\caption{\textsc{BuildHierarchy}: Recursive Hierarchical Clustering}
\label{alg:rhc}
\begin{algorithmic}[1]
\REQUIRE Embeddings $\mathcal{E}$, where $\mathcal{E} = \mathcal{E}_j$ for some node $j$, or $\mathcal{E} = \mathcal{E}_{\text{unk}}$; cluster size $k_{\max}$; large-corpus threshold $\tau$
\ENSURE  Binary tree root $r$
\STATE $k \leftarrow \min\!\bigl(k_{\max},\;\lfloor\sqrt{|\mathcal{E}|/2}\rfloor\bigr)$
\STATE $C \leftarrow \text{KMeans}(\mathcal{E},\; \text{max\_k}=k)$ \COMMENT{Top-down step}
\STATE $\{\mathcal{E}_{c}\}_{c} \leftarrow$ partition $\mathcal{E}$ by $C.\textit{kmeans\_preds}$
\STATE $r \leftarrow \text{Ward}(C.\textit{centers})$ \COMMENT{Bottom-up step}
\FOR{each leaf cluster $c$ in $r$}
    \IF{$|\mathcal{E}_{c}| > \tau$}
        \STATE $k' \leftarrow \min\!\bigl(k_{\max},\;\lfloor\sqrt{|\mathcal{E}_{c}|/2}\rfloor\bigr)$
        \STATE $\mathrm{sub} \leftarrow \textsc{BuildHierarchy}(\mathcal{E}_{c},\; k',\; \tau)$ \COMMENT{Recursion}
    \ELSE
        \STATE $\mathrm{sub} \leftarrow \text{Ward}(\mathcal{E}_{c})$ \COMMENT{Bottom-up step}
    \ENDIF
    \STATE Replace leaf node $c$ in $r$ with $\mathrm{sub}$
\ENDFOR
\RETURN $r$
\end{algorithmic}
\end{algorithm}

Given $\mathcal{E}_j$ for each taxonomy node $j$ and $\mathcal{E}_{\text{unk}}$, we construct the taxonomy by invoking Algorithm~\ref{alg:rhc} separately on each set. The algorithm produces a binary tree over the input embeddings by alternating between a top-down $k$-means partitioning step and a bottom-up Ward agglomerative merging step, applied recursively until every cluster is small enough for direct hierarchical clustering.

\stitle{Top-down step.}
Given an embedding set $\mathcal{E}$, the algorithm first partitions it into $K$ clusters via mini-batch $k$-means, where $K$ is chosen adaptively as
\[
    K \;=\; \min\!\left(K_{\max},\; \left\lfloor \sqrt{|\mathcal{E}|/2}\right\rfloor\right),
    \label{eq:adaptiveK}
\]
with $\lfloor \sqrt{|\mathcal{E}|/2} \rfloor$ being the standard rule-of-thumb estimate for the number of clusters~\cite{Multivariate1979} and $K_{\max}$ serving as an upper bound to prevent over-partitioning on large sets. This step yields $K$ centroids and an assignment of every embedding to its nearest centroid.

\stitle{Bottom-up step.}
Agglomerative clustering with Ward's minimum-variance linkage \cite{Ward01031963} is then applied to the $K$ centroids rather than to the raw embeddings. Operating on centroids keeps this step computationally tractable. Pairwise Ward dissimilarities between centroids are passed to the nearest-neighbor chain algorithm \cite{murtagh2011methods}, which efficiently produces a binary tree over the $K$ cluster regions.

\stitle{Recursion.}
Each leaf of this tree corresponds to a centroid that still represents many raw pseudo-labels. To resolve this, the algorithm recurses on the embeddings $\mathcal{E}_c$ assigned to each leaf cluster $c$: \myNum{i}~ If $|\mathcal{E}_c| > \tau$: the full top-down/bottom-up procedure is applied recursively, with $K_{\max}$ updated to $\min(K_{\max}, \lfloor\sqrt{|\mathcal{E}_c|/2}\rfloor)$, \myNum{ii}~ If $|\mathcal{E}_c| \leq \tau$: $k$-means is skipped and Ward linkage is applied directly to the raw embeddings, yielding the finest possible resolution at a scale where the full bottom-up sweep remains tractable. We set $\tau = 40{,}000$ in all experiments; the rationale for this choice is discussed in Section~\ref{sec:implementation}.


In each case, the resulting subtree replaces the corresponding leaf in the parent tree. Repeating this procedure for all leaves produces a fully resolved binary tree for each embedding set ($\mathcal{E}_j$, $\mathcal{E}_{\text{unk}}$).
The binary tree obtained from each $\mathcal{E}_j$ is then attached to its corresponding taxonomy node $j$, whereas the tree obtained from $\mathcal{E}_{\text{unk}}$ is attached directly to the taxonomy root.
Each internal node $v$ stores a tuple $(h(v), \mathrm{labels}(v))$, recording its Ward merge height, and the pseudo-labels assigned to the node $v$. Since the resulting tree is binary, whereas taxonomies are typically multi-way, we use these quantities in the following section to guide its conversion into a multi-way tree.




\subsection{Stage 4: Conversion}
\label{sec:method:contract}



At full depth, the constructed tree has as many leaves as there are pseudo-labels, which is too fine-grained. We therefore extract a set of \emph{effective leaves} by descending the tree and halting at any node $v$ whose Ward height satisfies $h(v) \le \delta_{\text{leaf}}$. In other words, every subtree whose root has a merge height below the threshold $\delta_{\text{leaf}}$ is treated as a single semantic cluster, and the pseudo-labels it contains are retained on that leaf for downstream naming. We set $\delta_{\text{leaf}} = 10$. Further discussion is provided in Section~\ref{sec:implementation}.

Additionally, as mentioned previously, the Recursive Hierarchical Clustering method produces strictly binary trees, but real taxonomies often have nodes with more than two children. We address this with \emph{height-tolerance collapsing}. Given a tolerance $\epsilon$ and an internal node $v$ with Ward merge height $h(v)$, we examine each child $u$ of $v$: if $u$ is an internal node and $h(v) - h(u) \le \epsilon$, we remove $u$ and lift its children to become direct children of $v$. We then recursively apply the same test to each lifted child, always comparing against the original height $h(v)$ rather than against the intermediate node's height. This anchoring is important: a chain of splits whose heights all lie within $\epsilon$ of $h(v)$ forms a single semantic plateau, and anchoring against $h(v)$ ensures the entire plateau collapses into one multi-way node at $v$. 
The procedure is applied top-down across the tree. 
We set $\epsilon = 20$; further discussion of this choice is provided in Section~\ref{sec:implementation}.


\subsection{Stage 5: LLM-Based Taxonomy Naming}
\label{sec:method:naming}

At this point, every newly created node in the taxonomy tree carries a list of raw pseudo-labels but lacks a human-readable name. We assign names via a \emph{bottom-up, position-aware} LLM summarization procedure, described in detail below. Note that nodes belonging to the predefined taxonomy already have names and are therefore excluded from this step.



\stitle{Leaf naming.}
For each leaf node, we prompt Qwen3-235B-A22B-2507 \cite{yang2025qwen3} with three inputs: \myNum{i}~Input Pseudo-Labels, containing all raw pseudo-labels assigned to the leaf; \myNum{ii}~Parent Path, containing the path of already-named ancestors (e.g., \emph{Privacy/Security $>$ Data Sharing}); and \myNum{iii}~Sibling Summaries, containing the names of sibling nodes processed earlier. We process sibling leaves sequentially so that the model can use previously generated sibling names when naming each subsequent leaf.

The model is instructed to generate a leaf name that \myNum{i}~contains $2$--$6$ words and is returned in JSON format, \myNum{ii}~represents a strict subcategory of the parent path, \myNum{iii}~is semantically distinct from the previously named siblings, and \myNum{iv}~avoids generic terms such as \emph{issues}, \emph{problems}, and \emph{concerns}. The complete leaf-naming prompt is provided in Table~\ref{tab:prompt-leaf}.

\stitle{Internal-node naming.}
After naming the leaf nodes, we name the remaining internal nodes bottom-up, processing each internal node only after all its direct children have been named. This order ensures that the child names are available when generating the name of their parent. We adapt the leaf-naming prompt by replacing Input Pseudo-Labels with the already-generated Child Names. The prompt requires the model to generate a parent name that \myNum{i}~encompasses all child names and is more general than each child, \myNum{ii}~sits properly under the parent path, and \myNum{iii}~is distinct from both the parent and sibling names.






\captionof{table}{Prompt for leaf node summarization. Template variables
\texttt{\$PARENT\_PATH\$}, \texttt{\$SIBLINGS\$}, and \texttt{\$LABELS\$} are
filled for each leaf node to be labeled.}
\begin{tcolorbox}[
    breakable,
    colback=gray!5,
    colframe=gray!50,
    title=Leaf Node Summarization Prompt,
    fonttitle=\bfseries,
    width=\columnwidth,
    fontupper=\footnotesize,
    boxsep=2pt,
    left=3pt,
    right=3pt,
    top=3pt,
    bottom=3pt
]
\raggedright
You are a precise taxonomy summarization assistant.
 
\smallskip
\textbf{Given:}
\begin{itemize}[noitemsep, topsep=2pt, parsep=0pt, leftmargin=*]
    \item A list of user review labels
    \item The parent path (hierarchical summaries above the current node) (if given)
    \item The sibling summaries (other summaries under the same parent) (if given)
\end{itemize}
 
Generate a concise subcategory summary that captures the dominant shared theme of the labels.
 
\smallskip
\textbf{Requirements:}
\begin{itemize}[noitemsep, topsep=2pt, parsep=0pt, leftmargin=*]
    \item The summary must be concise (2--6 words).
    \item The summary does not need to cover every label perfectly; prioritize the dominant theme.
    \item The summary must clearly be a subcategory of the provided parent path.
    \item The summary must not be the exact copy of parent summary or sibling summaries (if given).
    \item The summary must be at a comparable level of abstraction to the sibling summaries.
    \item The summary must be semantically distinct from its siblings (avoid overlap or near-duplicates).
    \item Prefer specific, concrete terms over generic ones. Avoid vague words (e.g., ``issues'', ``problems'', ``concerns'', ``practices'') unless the labels themselves are genuinely broad.
    \item Always output valid JSON with a single summary string.
    \item Do not include explanations or additional text.
\end{itemize}
 
\smallskip
\textbf{Output Format (JSON):}\\
\texttt{\{"summary": "\textless 2-6 word summary\textgreater"\}}
 
\smallskip
\textbf{Example:}
 
\textit{Parent Path:}\\
Privacy/Security $>$ Advertising
 
\smallskip
\textit{Sibling Summaries:}\\
Personalized advertising; Paid services
 
\smallskip
\textit{Input Labels:}\\
fake ads; scam advertisements; fraudulent promotions; misleading ads; phishing advertisements
 
\smallskip
\textit{Output:}\\
\texttt{\{"summary": "Fraudulent Ads"\}}
 
\smallskip
Now generate the concise summary using the following:
 
\smallskip
\textit{Parent Path:}\\
\texttt{[\$PARENT\_PATH\$]}
 
\smallskip
\textit{Sibling Summaries:}\\
\texttt{[\$SIBLINGS\$]}
 
\smallskip
\textit{Input Labels:}\\
\texttt{[\$LABELS\$]}
\end{tcolorbox}
\label{tab:prompt-leaf}
\vspace{-10pt}

\section{Experiments}

\subsection{Baselines}
\label{sec:baseline}

We compare {\ours} with four automatic taxonomy-construction methods. Because these methods differ in whether they require an initial taxonomy, we organize them into unseeded and seeded baselines. We evaluate {\ours} in both settings.

\textbf{Unseeded baselines.} Chain-of-Layers~\cite{ChainofLayer} constructs a taxonomy from scratch in a top-down, layer-by-layer manner. SCYCHIC~\cite{gao2025science} alternates between top-down embedding-based clustering and bottom-up higher-level abstraction.

\textbf{Seeded baselines.} TaxoCom~\cite{TaxoCom} recursively expands a given seed taxonomy by discovering subtopic clusters in a given corpus. TaxoAdapt~\cite{kargupta-etal-2025-taxoadapt} combines LLM knowledge with corpus signals via hierarchical classification. We evaluate it using both an LLM-generated seed and the expert taxonomy of Akgul et al.~\cite{privacyTaxonomy}.







Chain-of-Layers, TaxoAdapt, and SCYCHIC are constrained by the context window limitations of their underlying language models, making them unable to process large-scale pseudo-label sets directly. To accommodate these baselines, we apply k-means clustering to partition the pseudo-labels into 1,000 clusters and sample 100 labels from each cluster, yielding a representative subset of 100,000 samples that falls within their processing capacity. In contrast, {\ours} processes the complete corpus. To control for differences in data scale, we additionally evaluate {\ours} on the 100,000-label subset used by the baselines, denoting this variant as {\ours} (100K).

\subsection{Evaluation Metrics}

\stitle{Path Score:} For each parent-child pair, this metric uses an LLM to evaluate whether the child node is a genuinely more specific subtopic of its parent. We used GPT-4o~\cite{hurst2024gpt} to make this judgment. 

\stitle{Level Score:} For each parent node, this metric uses an LLM to evaluate how coherent and consistent in specificity its set of child nodes is relative to their parent. We used GPT-4o to make this judgment.

\stitle{Coverage Score:} For each pseudo-label in the corpus, this metric computes the maximum cosine similarity to any leaf node in the taxonomy. A pseudo-label is considered "covered" if its best similarity score meets or exceeds a threshold (default 0.6). The final score is the fraction of pseudo-labels that are covered, indicating how well the taxonomy's leaf nodes collectively account for the topics present in the corpus. A detailed comparison of coverage across multiple similarity thresholds is presented in Section~\ref{sec:coverage}.

\stitle{Novelty Score:} For each node in the taxonomy, this metric uses an LLM to evaluate whether the node represents a novel and meaningful concept beyond what a standard Privacy/Security taxonomy would include.
The final reported score is a count-weighted average across all nodes, where each node's score is weighted by how many labels it covers, giving more influence to nodes that represent larger portions of the corpus. We again used GPT-4o to make this judgment.

We evaluate the reliability of the LLM-based metrics through a human validation study, as described in Section~\ref{sec:human-validation}.

\begin{table*}[ht]
\centering
\caption{Comparison of taxonomy methods. \textbf{Bold} is the best score per seeding regime, \underline{underline} the second best.}
\label{tab:taxonomy_comparison}
\setlength{\tabcolsep}{10pt}
\resizebox{0.7\textwidth}{!}{%
\begin{tabular}{llcccc}
\toprule
\textbf{Method} & \textbf{Seed} & \textbf{Path} & \textbf{Level} & \textbf{Coverage} & \textbf{Novelty} \\
\midrule
Chain of Layers & --     & \underline{0.6533} & 0.4444 & \underline{0.9090} & 0.1092 \\
SCYCHIC         & --     & 0.4652 & \underline{0.6232} & 0.5493 & \textbf{0.4219} \\
{\ours} (Ours)  & --     & \textbf{0.7980} & \textbf{0.7258} & \textbf{0.9537} & \underline{0.1933} \\
\midrule
TaxoCom         & Akgul et al.~\cite{privacyTaxonomy} & 0.4830 & 0.4285 & 0.8976 & 0.1213 \\
TaxoAdapt       & LLM Generated   & \underline{0.7031} & 0.7407 & 0.5671 & 0.2788 \\
TaxoAdapt       & Akgul et al.~\cite{privacyTaxonomy} & 0.6785 & 0.7200 & 0.8938 & 0.2387 \\
{\ours} (Ours)  & Akgul et al.~\cite{privacyTaxonomy} & \textbf{0.7194} & \textbf{0.7672} & \textbf{0.9854} & \textbf{0.4346} \\
{\ours} (100K)
& Akgul et al.~\cite{privacyTaxonomy}
& 0.6649
& \underline{0.7642}
& \underline{0.9677}
& \underline{0.4105} \\
\bottomrule
\end{tabular}
}
\end{table*}


\begin{table}[t]
\caption{runtime and peak memory of Ward agglomerative as a function of subcluster size $n$, measured on our machine (Core i9-9920X CPU) with embedding dimension $d = 4096$.} 
\centering
\small
\begin{tabular}{rrr}
\toprule
$\mathbf{n}$ & \textbf{Runtime} & \textbf{$\Delta$RAM (MB)} \\
\midrule
5{,}000  & 33s          & 253 \\
10{,}000 & 2m 16s       & 763 \\
20{,}000 & 9m 15s       & 3{,}052 \\
40{,}000 & 37m 28s      & 13{,}457 \\
60{,}000 & 1h 24m       & 29{,}340 \\
\bottomrule
\end{tabular}
\label{tab:tau-scaling}
\vspace{-10pt}
\end{table}


\subsection{Results and Analysis}

As shown in Table~\ref{tab:taxonomy_comparison}, {\ours} performs strongly under both seeding settings. Without an initial taxonomy, {\ours} achieves the highest path (0.7980), level (0.7258), and coverage (0.9537) scores among the unseeded methods. These results demonstrate that the effectiveness of our approach does not depend solely on an expert-provided taxonomy: Recursive Hierarchical Clustering can independently recover coherent parent--child relationships and maintain consistent granularity across levels.
Although SCYCHIC obtains higher novelty in the unseeded setting, its substantially lower path, level, and coverage scores indicate a trade-off between discovering unconventional concepts and producing a coherent, representative taxonomy.

When initialized with the taxonomy proposed by Akgul et al.~\cite{privacyTaxonomy}, {\ours} achieves the best performance across all four metrics within the seeded regime. It obtains a path score of 0.7194 and a level score of 0.7672, surpassing the strongest corresponding baseline, TaxoAdapt with an LLM-generated seed. Its coverage score of 0.9854 also substantially exceeds the strongest seeded baseline, TaxoCom (0.8976), showing that {\ours} can extend an existing taxonomy while accounting for nearly all topics represented in the large-scale pseudo-label corpus. Moreover, {\ours} achieves a weighted novelty score of 0.4346, improving over the best seeded baseline by 0.1558 and even surpassing the strongest unseeded result from SCYCHIC (0.4219). This finding suggests that grounding the hierarchy in an existing taxonomy does not prevent the discovery of new concepts; instead, the seed provides a stable semantic structure from which corpus-specific categories can be introduced.

Comparing the unseeded and seeded variants of {\ours} highlights the effect of incorporating the seed taxonomy proposed by Akgul et al.~\cite{privacyTaxonomy}. Seeding improves the level score from 0.7258 to 0.7672, coverage from 0.9537 to 0.9854, and weighted novelty from 0.1933 to 0.4346. 
These results indicate that the expert taxonomy provides benefits for level consistency, coverage, and novelty, although with a trade-off in parent--child path quality. Overall, the results show that {\ours} remains competitive without prior structure while benefiting from expert knowledge when a seed taxonomy is available.

The matched-budget variant, \ours{} (100K), enables us to separate the effect of our method from that of corpus scale. Even under this identical data budget, {\ours} remains the strongest method on three of the four metrics: it achieves a level score of 0.7642, a coverage of 0.9677, and a weighted novelty of 0.4105, exceeding the best corresponding baseline results. These results indicate that {\ours}'s advantages do not stem merely from access to more data. At the same time, the advantage of full {\ours} over {\ours} (100K) demonstrates the benefit of processing the complete corpus.



\begin{table*}[ht]
\caption{Coverage scores at five similarity thresholds.
\textbf{Bold} indicates the best score per seeding regime, and \underline{underline} indicates the second best.}
\label{tab:coverage}
\centering
\resizebox{0.7\textwidth}{!}{%
\begin{tabular}{llrrrrr}
\toprule
\textbf{Method}
& \textbf{Seed}
& \textbf{0.5}
& \textbf{0.6}
& \textbf{0.7}
& \textbf{0.8}
& \textbf{0.9} \\
\midrule
Chain of Layers
& -- 
& \underline{0.9934}
& \underline{0.9090}
& \underline{0.5298}
& \underline{0.1437}
& \underline{0.0092} \\

SCYCHIC
& --
& 0.9475
& 0.5493
& 0.0717
& 0.0010
& 0.0000 \\

{\ours} (Ours)
& --
& \textbf{0.9967}
& \textbf{0.9537}
& \textbf{0.6784}
& \textbf{0.1836}
& \textbf{0.0096} \\
\midrule
TaxoCom
& Akgul et al.~\cite{privacyTaxonomy}
& 0.9905
& 0.8976
& 0.4126
& 0.0634
& 0.0027 \\

TaxoAdapt
& LLM Generated
& 0.9490
& 0.5671
& 0.1556
& 0.0131
& 0.0001 \\

TaxoAdapt
& Akgul et al.~\cite{privacyTaxonomy}
& 0.9919
& 0.8938
& 0.4103
& 0.0534
& 0.0022 \\

{\ours} (Ours)
& Akgul et al.~\cite{privacyTaxonomy}
& \textbf{0.9988}
& \textbf{0.9854}
& \textbf{0.8540}
& \textbf{0.3725}
& \textbf{0.0297} \\

{\ours} (100k)
& Akgul et al.~\cite{privacyTaxonomy}
& \underline{0.9976}
& \underline{0.9677}
& \underline{0.7318}
& \underline{0.2099}
& \underline{0.0140} \\

\bottomrule
\end{tabular}%
}
\vspace{-10pt}
\end{table*}

\subsection{Scalability and Hyperparameter Choices}
\label{sec:implementation}

Direct Ward clustering becomes increasingly expensive as the subcluster size grows. Its nearest-neighbor-chain implementation has $O(n^2)$ time complexity~\cite{murtagh2011methods}, and our implementation also exhibits quadratic memory growth. As shown in Table~\ref{tab:tau-scaling}, its runtime and memory usage increase sharply beyond 40{,}000 points, reaching 1h\,24m and 29~GB at 60{,}000 points. We therefore set $\tau=40{,}000$: subclusters at or below this threshold are processed directly with Ward, whereas larger subclusters are first partitioned using k-means.

This threshold enables Recursive Hierarchical Clustering and Conversion (Stages~3 and~4) to process all 512{,}520 unique pseudo-label embeddings in 17m\,19s, with peak memory below 8~GB, on a single Intel Core i9-9920X CPU. In comparison, applying Ward directly to the complete corpus would be infeasible on the same machine.



As it was mentioned previously, we set $\delta_{\text{leaf}} = 10$ and $\epsilon = 20$. The leaf height threshold $\delta_{\text{leaf}}$ controls how deep we walk the binary tree before treating a sub-tree as a single semantic cluster: lower values produce finer, more specific leaves, while higher values produce broader leaves containing more diverse pseudo-labels. We chose a relatively low value so that each finalized leaf corresponds to a tight, coherent theme, which is important for Step 5: the LLM namer conditions on the pseudo-labels assigned to a leaf, and clusters spanning semantically distant pseudo-labels would force the model toward overly generic names. At the same time, $\delta_{\text{leaf}}$ cannot be arbitrarily small; values close to zero split nearly identical pseudo-labels (e.g., ``tracks my location'' and ``my location is being tracked'') into separate leaves, producing redundant siblings. $\delta_{\text{leaf}} = 10$ balances these concerns. 
Moreover, the height-tolerance $\epsilon$ governs how aggressively near-equal-height splits are collapsed into multi-way nodes: low values preserve the binary structure with narrow two-way splits, while high values lift many children to their grandparents, producing flatter nodes that may group children of differing specificity under a single parent. In our sweep, weighted novelty peaks at $\epsilon = 20$ and decreases for both smaller and larger values. Beyond $\epsilon = 20$, we also observe a sharp drop in path granularity, as children of unequal specificity are forced under the same parent. We therefore set $\epsilon = 20$. 

\subsection{Coverage Across Similarity Thresholds}
\label{sec:coverage}

For a more comprehensive evaluation, we report coverage scores for all baselines across a range of similarity thresholds $\in \{0.5, 0.6, 0.7, 0.8, 0.9\}$. As shown in Table~\ref{tab:coverage}, the unseeded and seeded variants of {\ours} achieve the highest coverage within their respective seeding regimes at every threshold. The performance difference is most pronounced at a threshold of 0.7, where the seeded variant achieves 0.8540 coverage compared with 0.5298 for the strongest baseline, an absolute improvement of 0.3242. The unseeded variant also achieves 0.6784 at this threshold, outperforming the strongest unseeded baseline by 0.1486. Notably, even when restricted to the baselines' 100K-sample budget, {\ours} (100K) attains 0.7318 coverage at this threshold, exceeding the strongest baseline by 0.2020, which confirms that the coverage advantage is attributable to the method rather than to access to the full corpus. Together, these results demonstrate that the coverage advantage of {\ours} persists across similarity thresholds and depends on neither the availability of a seed taxonomy nor a larger data budget.

\begin{figure*}[t]
    \centering
    \begin{minipage}{0.49\textwidth}
        \centering
        \includegraphics[width=\textwidth]{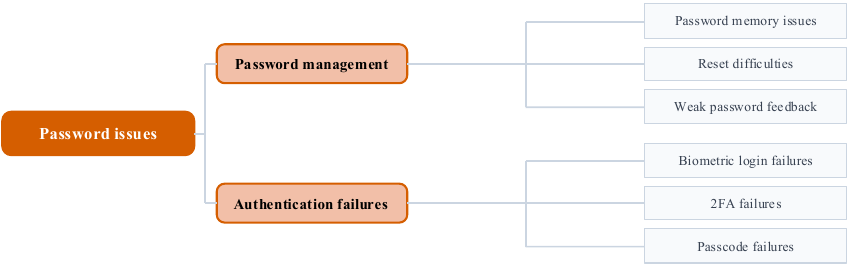}
        \caption{Password Branch}
        \label{fig:password_tax}
    \end{minipage}\hspace{0.01\textwidth}
    \begin{minipage}{0.49\textwidth}
        \centering
        \includegraphics[width=\textwidth]{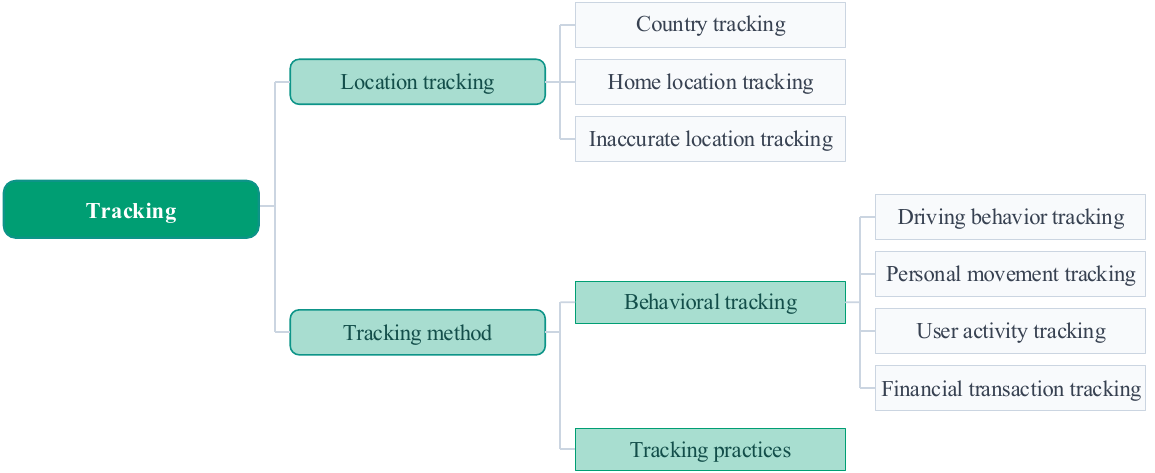}
        \caption{Location \& Tracking Branch}
        \label{fig:location_tax}
    \end{minipage}
\end{figure*}

\begin{figure*}[t]
    \centering

    \begin{minipage}{0.49\textwidth}
        \centering
        \includegraphics[width=\textwidth]{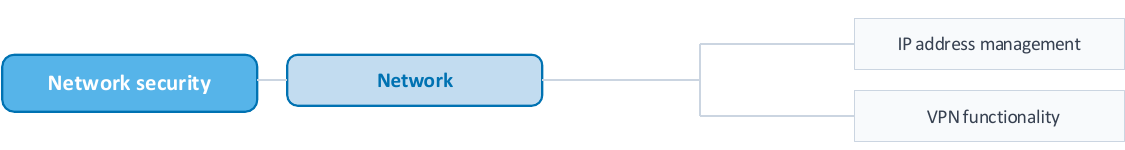}
        \caption{Network Branch}
        \label{fig:network_tax}
    \end{minipage}\hspace{0.01\textwidth}
    \begin{minipage}{0.49\textwidth}
        \centering
        \includegraphics[width=\textwidth]{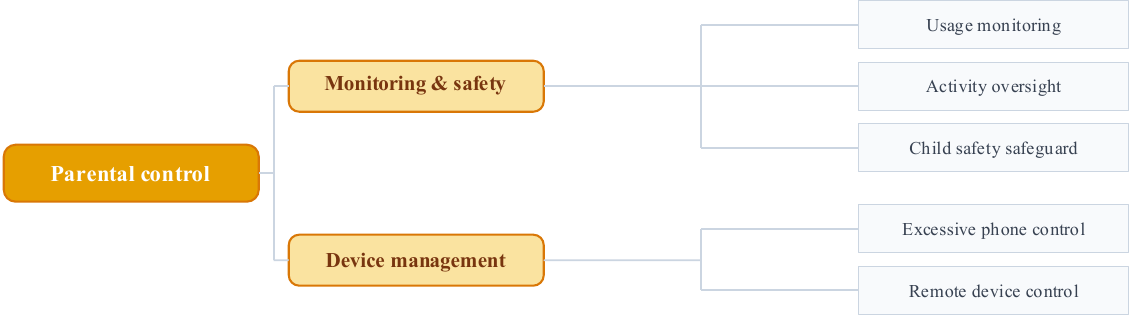}
        \caption{Parental Control Branch}
        \label{fig:parental_tax}
    \end{minipage}
    \vspace{10pt}
\end{figure*}

\subsection{Human Validation of LLM-Based Metrics}
\label{sec:human-validation}

To validate the LLM-based evaluation metrics, we conducted a human annotation study. For each of the three LLM-judged Path Score, Level Score, and Novelty, we sampled 100 items from the generated taxonomies. Two annotators independently labeled each item following a written annotation guide, using the same scales as the LLM judge. We report inter-annotator agreement and agreement between GPT-4o and each annotator (Cohen's $\kappa$ \cite{Cohen1960ACO}). As a complementary view, we additionally report GPT-4o's agreement on the subset of items where both annotators independently assigned the same label, i.e., the unambiguous cases.

\stitle{Path Score:} For parent–child granularity, GPT-4o achieves moderate agreement with each annotator ($\kappa$ = 0.459 and $\kappa$ = 0.461; 74\% and 72\% raw agreement), exceeding the inter-annotator agreement itself ($\kappa$ = 0.351; 66\%), which indicates that the LLM judge falls within the range of variation between human annotators. On the 66 items where both annotators independently agreed, GPT-4o matches the unanimous human label in 84.8\% of cases ($\kappa$ = 0.69), corresponding to substantial agreement. These results support the reliability of the Path Score.

\stitle{Level Score:} Level granularity is rated on an ordinal 0--3 scale, so we report quadratic-weighted $\kappa$, which gives partial credit when ratings are close (e.g., 2 vs.\ 3) rather than treating every mismatch as equal. By this measure, GPT-4o agrees with Annotator 1 at $\kappa$ = 0.465, higher than the agreement between the two annotators themselves ($\kappa$ = 0.250). On the 35 items where both annotators independently gave the same score, agreement rises to $\kappa$ = 0.503. We also binarize the four-level scale, grouping scores of 0--1 as inadequate granularity and scores of 2--3 as adequate granularity. Here GPT-4o matches Annotator 1 on 81\% of items, and on the 53 items where both annotators agree under the binarized scale, GPT-4o matches their shared label on 79.2\% ($\kappa$ = 0.404). Overall, the LLM judge tracks human granularity judgments at the ordinal level with moderate agreement.

\stitle{Novelty Score:} Novelty proves to be the most subjective dimension: even the two human annotators reach only slight agreement ($\kappa$ = 0.189 despite 74\% raw agreement), with markedly different base rates (27\% vs. 11\% of nodes labeled novel), and GPT-4o’s agreement with each annotator ($\kappa$ = 0.174 and 0.228) is comparable to the agreement observed between the two human annotators. The model applies a systematically more liberal novelty criterion, labeling 44\% of nodes as novel. 
Given the low human ceiling and the judge's uniform leniency across all compared methods, we interpret Weighted Novelty strictly as a relative measure for comparing methods under a consistently applied criterion, rather than as an absolute, human-aligned assessment of novelty; we complement it with the qualitative analysis of novel branches in the next section (\ref{sec:novel_nodes}). 

\vspace{-2pt}
\subsection{Qualitative Analysis of Novel Nodes}
\label{sec:novel_nodes}

{\ours} generates taxonomies that surface interesting and novel branches. For example, in Figure \ref{fig:location_tax}, the Behavioral Tracking node expands into four meaningful subcategories: Driving Behavior Tracking, Personal Movement Tracking, User Activity Tracking, and Financial Transaction Tracking; each capturing a distinct dimension of user surveillance. Similarly, Figure \ref{fig:password_tax} reveals nuanced nodes such as Weak Password Feedback, Reset Difficulties, and 2FA Failures.

Beyond refining existing branches, {\ours} uncovers a Network Security branch not present in the original privacy/security taxonomy (Figure~\ref{fig:network_tax}), with subcategories such as IP Address Management and VPN Functionality. Furthermore, because our taxonomy is constructed from reviews of Teacher Approved apps alongside other datasets, it captures a domain-specific branch absent from prior work; Parental Control (Figure \ref{fig:parental_tax}), with nodes such as Child Safety Safeguards and Activity Oversight.

\section{Related work}

\stitle{Hand-crafted Privacy/Security Taxonomies.} 
Prior work has explored automated analysis of Google Play app reviews through the lens of privacy and security. 
Nema et al.\cite{Security_Privacy_Nema} developed an NLP-based methodology to classify and cluster privacy-related reviews, and Harkous et al.\cite{Security_Privacy_Harkous} introduced Hark, an end-to-end deep learning system for discovering and summarizing privacy-related feedback from app reviews. Neither of these methods proposed a privacy taxonomy grounded in the review data itself.
A subsequent study analyzed privacy-relevant reviews spanning a decade, and proposed a hand-crafted privacy taxonomy \cite{privacyTaxonomy}. More broadly, foundational privacy taxonomies have also been proposed outside the app review context \cite{solove2006taxonomy, Privacy_enhancing_Wang}. In all cases, these taxonomies were constructed manually and were not designed for the evolving, user-generated nature of app review data.


\stitle{Automatic Taxonomy Creation with LLMs.}
To address the scalability limitations of manual taxonomy construction, researchers have proposed LLM-based methods for automatic taxonomy generation. Several works have explored using LLMs to iteratively generate and refine label taxonomies from unstructured text~\cite{TnT_Wan, wang-etal-2023-goal, ChainofLayer, hsu-etal-2024-chime}. TaxoAdapt~\cite{kargupta-etal-2025-taxoadapt} further advances this direction by dynamically adapting LLM-generated taxonomies to an evolving corpus across multiple dimensions. The problem with these methods is that they rely primarily on LLM prompting and reasoning to induce structure and do not take advantage of traditional embedding-based clustering.


\stitle{Automatic creation with clustering.}
Another line of work constructs topic taxonomies by grounding structure in embedding geometry. TaxoGen~\cite{TaxoGen_Zhang} learns term embeddings and recursively applies spherical clustering to split coarse topics into finer ones. NetTaxo~\cite{NetTaxo_Shang} extends this to text-rich networks, learning term embeddings from both text and network structure. TaxoCom~\cite{TaxoCom} uses discriminative embeddings and novelty adaptive clustering to expand a partial hierarchy. These methods rely purely on corpus-derived embeddings and do not leverage the knowledge encoded in large language models.





\stitle{Automatic creation with LLM and clustering}
A recent line of work combines the knowledge of LLMs with embedding-based clustering~\cite{hu2024taxonomy, katz-etal-2024-knowledge, gao2025science, zhu-etal-2025-context}. Like ours, these methods integrate LLMs with clustering; however, they are designed for organizing scientific literature and are evaluated on corpora of at most around ten thousand documents. In contrast, our Recursive Hierarchical Clustering method scales to our dataset of 600K noisy app reviews.






\section{Conclusion}
We presented {\ours}, a five-step pipeline that automatically expands an expert-defined privacy and security taxonomy from large-scale mobile app reviews. By combining \emph{Recursive Hierarchical Clustering} with LLM-based node naming, {\ours} is applied to 601,257 privacy- and security-related reviews and outperforms strong automatic-taxonomy baselines on path, level, coverage, and novelty metrics. Beyond refining existing categories, it surfaces concrete novel branches absent from prior taxonomies. We released the filtered review corpus, extracted concern labels, and the {\ours} codebase as a reusable resource for future privacy and security research on mobile apps.


\bibliographystyle{IEEEtran}

\bibliography{sample-base.bib}

\end{document}